\documentclass[10pt,twocolumn,letterpaper]{article}

\usepackage[pagenumbers]{cvpr} 

\usepackage[dvipsnames]{xcolor}

\definecolor{cvprblue}{rgb}{0.21,0.49,0.74}
\usepackage[pagebackref,breaklinks,colorlinks,citecolor=cvprblue]{hyperref}
\usepackage{multirow}
\usepackage{array}
\usepackage{booktabs}
\usepackage{algorithm}
\usepackage{algpseudocode}
\usepackage{amsfonts}

\newcolumntype{L}[1]{>{\raggedright\arraybackslash}m{#1}}
\newcolumntype{C}[1]{>{\centering\arraybackslash}m{#1}}
\newcolumntype{R}[1]{>{\raggedleft\arraybackslash}m{#1}}
\newcolumntype{+}{>{\global\let\currentrowstyle\relax}}
\newcolumntype{^}{>{\currentrowstyle}}

\title{Clicks2Line: Using Lines for Interactive Image Segmentation}

\author{Chaewon Lee\\
Korea University\\
{\tt\small chaewonlee@mcl.korea.ac.kr}
\and
Chang-Su Kim\\
Korea University\\
{\tt\small changsukim@korea.ac.kr}
}

\begin{document}
\maketitle
\begin{abstract}
For click-based interactive segmentation methods, reducing the number of clicks required to obtain a desired segmentation result is essential. Although recent click-based methods yield decent segmentation results, we observe that substantial amount of clicks are required to segment elongated regions. To reduce the amount of user-effort required, we propose using lines instead of clicks for such cases. In this paper, an interactive segmentation algorithm which adaptively adopts either clicks or lines as input is proposed. Experimental results demonstrate that using lines can generate better segmentation results than clicks for several cases.
\end{abstract}

\section{Introduction}
\label{sec:intro}

Interactive image segmentation is a task that aims to segment objects of interest given guidance through user annotations. Various forms of user-annotations have been introduced including clicks \cite{jang2019interactive, sofiiuk2020f, sofiiuk2022reviving,liu2023simpleclick,lee2024interactive}, scribbles \cite{grady2006random, bai2014error} and boxes \cite{lempitsky2009image,rother2004grabcut}. However, click-based methods have become the mainstream methods due to their simplicity and well-established protocols.

\begin{figure}[t]
    \begin{center}
    \includegraphics[width=\linewidth]{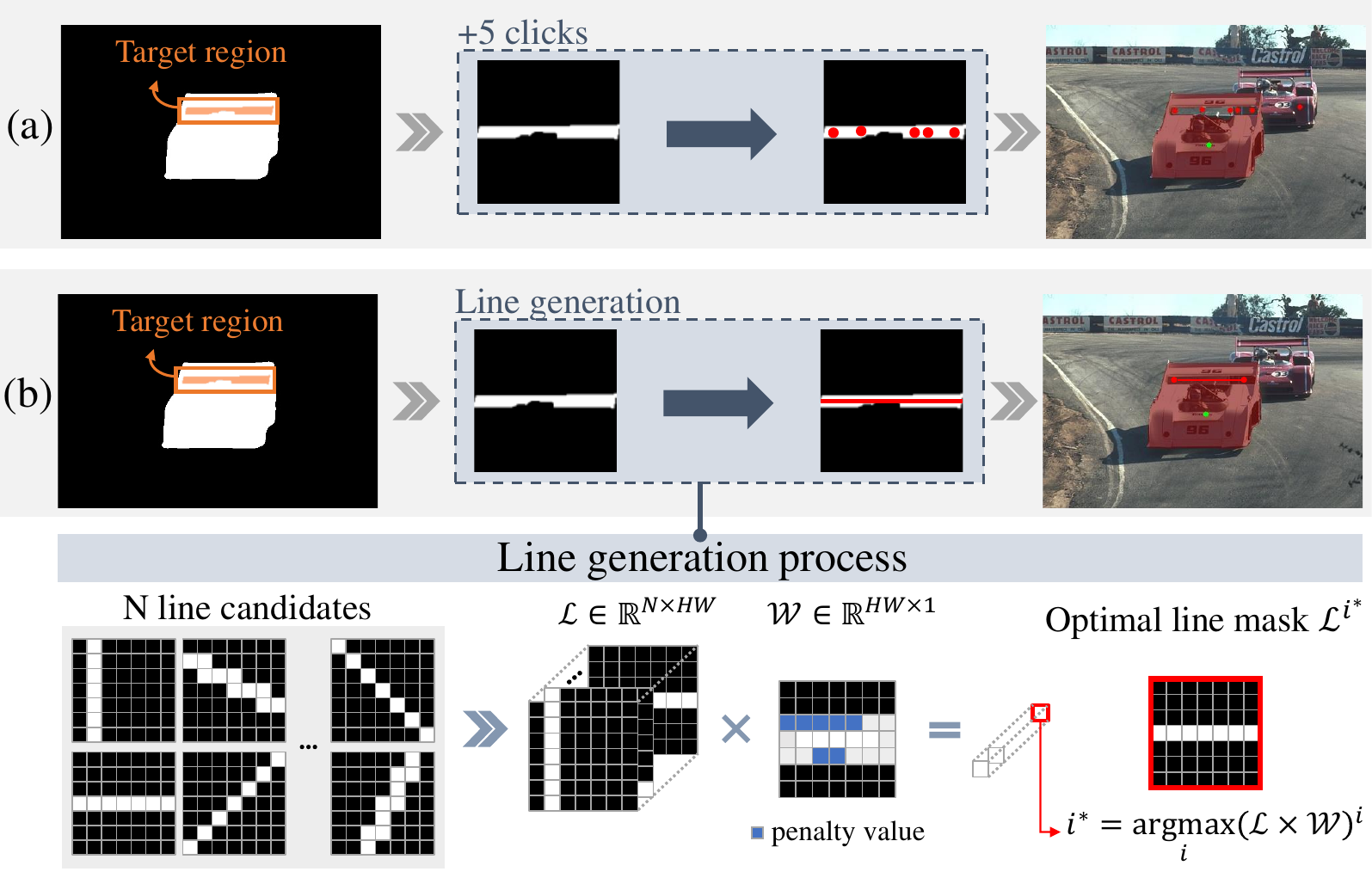}
    \end{center}
    \vspace*{-0.5cm}
    \caption
    {
        Comparison of existing algorithm and the proposed algorithm. Row (a) shows the input needed to segment the target region for the SimpleClick\cite{liu2023simpleclick} method, while row (b) shows that of the proposed method. The line generation process of the proposed method is also illustrated below.
    }
    \label{fig:line_generation}
\end{figure}

The objective of interactive image segmentation is to obtain high-quality segmentation results with less amount of user-annotations. Although existing click-based methods yield decent segmentation results, they require substantial amount of user clicks to segment long regions as in the case of Figure~\ref{fig:line_generation}(a). Generally, a line can cover more pixels than a click, thus it provides much better representations of elongated objects. Therefore, one could expect that a single line, as in Figure~\ref{fig:line_generation}(b), could better indicate a target's appearance than multiple clicks. Based on this observation, we propose using lines instead of clicks to extract segmentation results in such cases. In this paper, we propose an interactive image segmentation algorithm, called Clicks2Line, which adaptively adopts either clicks or lines as user-input. Experimental results show that the proposed algorithm successfully segments elongated regions with less user-annotations.

\section{Proposed Algorithm}
\label{sec:proposed_algorithm}

\subsection{Input selection: click vs. line}
A line can be defined using two points, thus we generate a line by connecting two clicks. Although lines are beneficial for targeting long objects, it requires two clicks. Therefore, for rather ordinary cases(\eg non-elongated objects), providing a click seems to be more advantageous as it requires less number of clicks. We construct our algorithm so that the model could adaptively take either a click or a line as input according to the target's shape. If the target region that the user wants to refine has an aspect ratio smaller than $q$, we employ a click as input. If the aspect ratio is greather than or equal to $q$, then we employ a line as input.

\begin{figure*}[t]
    \begin{center}
    \includegraphics[width=\linewidth]{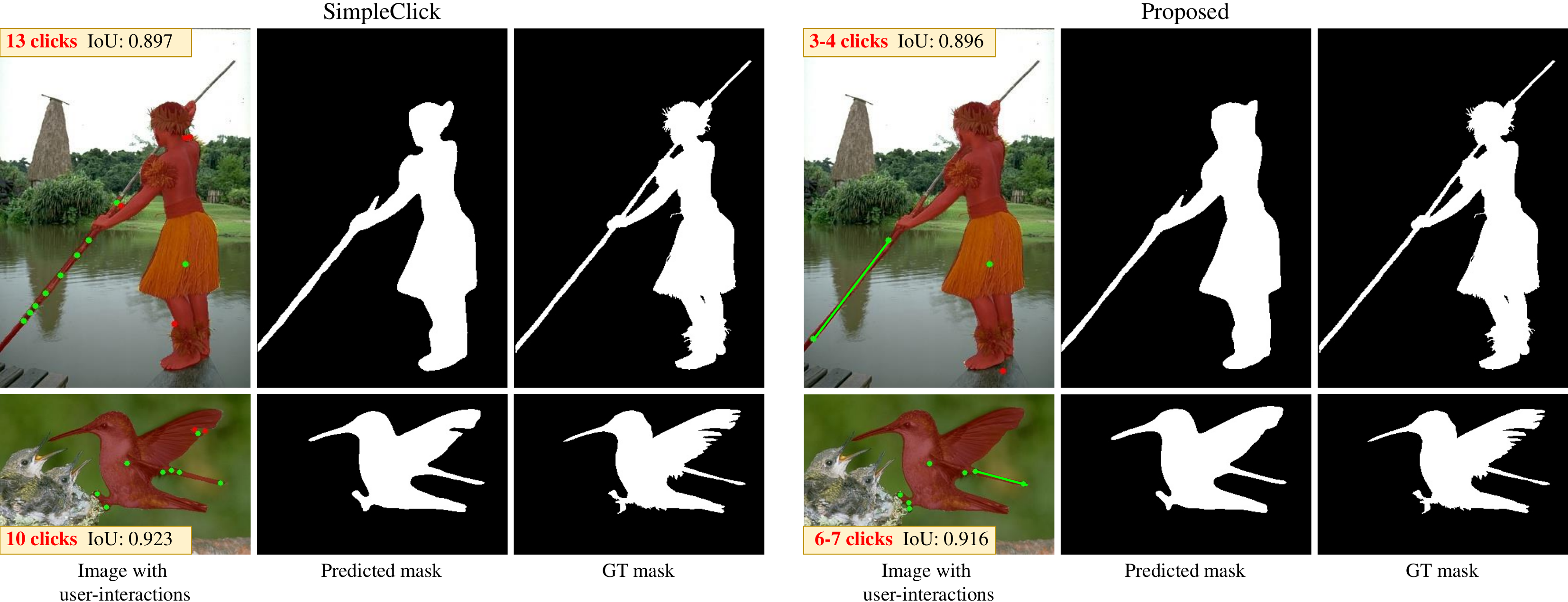}
    \end{center}
    \vspace*{-0.5cm}
    \caption
    {
        Qualitative comparison of algorithms
    }
    \label{fig:qualitative_comparison}
    \vspace*{-0.25cm}
\end{figure*}

\subsection{Line generation}
\label{sec:line_generation}
In order to draw a line that best represents the target region, it seems natural for the line to satisfy the following traits. 1) The line should be the longest line representing the target region. 2) The line should cross only the regions that it's class label belongs to (\eg a foreground/background line should only cross foreground/background regions).

To realize this, we generate a line through the following procedure. First, $N$ line candidates are generated to form a line candidates map $\mathcal{L}=[\mathcal{L}^1, \mathcal{L}^2, ..., \mathcal{L}^N]\in \mathbb{R}^{N \times H \times W}$. Next, we obtain target map $\mathcal{T}$ by cropping around the bounding box of the target region. The target map is padded and resized to $H \times W$, the same resolution as a line candidate map. According to the target map, we construct a weight map $\mathcal{W} \in \mathbb{R}^{H \times W}$ that gives more positive weight if the pixel is closer to the target region's center, and assigns big negative values if the pixel does not belong to the same class as the target. The negative weight value is used for the purpose of penalizing lines that cross opposite label regions. The negative weight value is fixed to $k$ for the experiments in this paper. Out of the $N$ line candidates from $\mathcal{L}$, the most optimal line map $\mathcal{L}^{i^*}$ is selected by

\begin{equation}\label{eq:optimal_line}
    i^{*} = \arg \max_{i} ({\mathcal{L} \times \mathcal{W}})^{i}
\end{equation}

Note that before the matrix multiplication between $\mathcal{L}$ and $\mathcal{W}$ is performed, $\mathcal{L}$ is reshaped to $\mathbb{R}^{N \times HW}$ and $\mathcal{W}$ to $\mathbb{R}^{HW \times 1}$. Once $\mathcal{L}^{i^*}$ is obtained, we compute the intersection of $\mathcal{L}^{i^*}$ and $\mathcal{T}$ and determine the endpoints of the intersected line as the two clicks consisting the input line.

\section{Experiments}
\label{sec:experiments}

\subsection{Implementation Details}
We adopt the network architecture in \cite{lee2024interactive} and take ViT-B \cite{dosovitskiy2021an}as the backbone network. For training, we use the SBD dataset \cite{hariharan2011semantic}.We minimize the normalized focal loss \cite{sofiiuk2022reviving} using the Adam optimizer with $\beta_1=0.9$ and $\beta_2=0.999$. The network is trained for 35 epochs. For evaluation, we use the GrabCut \cite{rother2004grabcut}and Berkeley \cite{mcguinness2010comparative} datasets.

When determining the input type, we always use a click for the first round. On later rounds, the aspect ratio $q$ for determining whether to input a click or a line is fixed to 5. If the input is a click, we adopt the automatic clicking strategy used in \cite{xu2016deep, jang2019interactive} to generate a click. If the input is a line, we generate a line according to the process explained in Section ~\ref{sec:line_generation}.

\subsection{Performance Assessment}

\noindent\textbf{Quantitative Results:} Since the input line is generated by connecting two clicks, we could assess the proposed Clicks2Line using the NoC metrics. We compare Clicks2Line with a few conventional click-based interactive segmentation algorithms trained on the SBD dataset. For fair comparison, we only compare algorithms that use the same backbone network, ViT-B \cite{dosovitskiy2021an}. Table ~\ref{table:comparative_assessment_sbd} lists the results. Except for NoC85, the proposed algorithm shows the best results for the remaining tests.

\noindent\textbf{Qualitative Results:} Figure ~\ref{fig:qualitative_comparison} compares qualitative segmentation results of the proposed algorithm with that of SimpleClick \cite{liu2023simpleclick}. It shows that the proposed algorithm requires less user-clicks to segment elongated regions.

\begin{table}[t]\centering
    \vspace*{-0.2cm}
    \renewcommand{\arraystretch}{1.2}
    \caption
    {
        Evaluation results on the GrabCut and Berkeley datasets.
    }
    \vspace*{-0.15cm}
    \resizebox{1.0\linewidth}{!}{
    \begin{tabular}[t]{+L{3.0cm}^C{1.0cm}^C{1.0cm}^C{1.0cm}^C{1.0cm}^C{1.0cm}^C{1.0cm}}
    \toprule
     \multirow{2}[2]{*}{} & \multicolumn{3}{c}{GrabCut} &  \multicolumn{3}{c}{Berkeley}\\
     \cmidrule(lr){2-4} \cmidrule(lr){5-7}
    Algorithm & NoC85 & NoC90& NoC95& NoC85 & NoC90& NoC95\\
    \midrule
         SimpleClick \cite{liu2023simpleclick} & \underline{1.40} & 1.54 & 2.16 & \underline{1.44} & 2.46 & 6.70 \\
         MFP \cite{lee2024interactive} & \bf{1.38} & \underline{1.48} & \underline{1.92}  & \bf{1.39} &\underline{2.17} & \underline{6.18} \\
    \midrule
         Clicks2Line {\scriptsize (Proposed)} & 1.44 & \bf{1.46} & \bf{1.88}  & \underline{1.44} &\bf{2.04} & \bf{5.98} \\
    \bottomrule
    \end{tabular}}
    \label{table:comparative_assessment_sbd}
    \vspace*{-0.1cm}
\end{table}

\section{Conclusions}
\label{sec:conclusion}
In this paper, we proposed the Clicks2Line algorithm, which adaptively takes either clicks or lines as input according to the target's appearance. We showed that the algorithm is more advantageous especially for situations where the target object is long.

{
    \small
    \bibliographystyle{ieeenat_fullname}
    \bibliography{main}

\begin{thebibliography}{13}
\providecommand{\natexlab}[1]{#1}
\providecommand{\url}[1]{\texttt{#1}}
\expandafter\ifx\csname urlstyle\endcsname\relax
  \providecommand{\doi}[1]{doi: #1}\else
  \providecommand{\doi}{doi: \begingroup \urlstyle{rm}\Url}\fi

\bibitem[Bai and Wu(2014)]{bai2014error}
Junjie Bai and Xiaodong Wu.
\newblock Error-tolerant scribbles based interactive image segmentation.
\newblock In \emph{CVPR}, pages 392--399, 2014.

\bibitem[Dosovitskiy et~al.(2021)Dosovitskiy, Beyer, Kolesnikov, Weissenborn,
  Zhai, Unterthiner, Dehghani, Minderer, Heigold, Gelly, Uszkoreit, and
  Houlsby]{dosovitskiy2021an}
Alexey Dosovitskiy, Lucas Beyer, Alexander Kolesnikov, Dirk Weissenborn,
  Xiaohua Zhai, Thomas Unterthiner, Mostafa Dehghani, Matthias Minderer, Georg
  Heigold, Sylvain Gelly, Jakob Uszkoreit, and Neil Houlsby.
\newblock An image is worth 16x16 words: Transformers for image recognition at
  scale.
\newblock In \emph{ICLR}, 2021.

\bibitem[Grady(2006)]{grady2006random}
Leo Grady.
\newblock Random walks for image segmentation.
\newblock \emph{IEEE TPAMI}, 28\penalty0 (11):\penalty0 1768--1783, 2006.

\bibitem[Hariharan et~al.(2011)Hariharan, Arbel{\'a}ez, Bourdev, Maji, and
  Malik]{hariharan2011semantic}
Bharath Hariharan, Pablo Arbel{\'a}ez, Lubomir Bourdev, Subhransu Maji, and
  Jitendra Malik.
\newblock Semantic contours from inverse detectors.
\newblock In \emph{ICCV}, pages 991--998, 2011.

\bibitem[Jang and Kim(2019)]{jang2019interactive}
Won-Dong Jang and Chang-Su Kim.
\newblock Interactive image segmentation via backpropagating refinement scheme.
\newblock In \emph{CVPR}, pages 5297--5306, 2019.

\bibitem[Lee et~al.(2024)Lee, Lee, and Kim]{lee2024interactive}
Chaewon Lee, Seon-Ho Lee, and Chang-Su Kim.
\newblock {MFP}: Making full use of probability maps for interactive image
  segmentation.
\newblock In \emph{CVPR}, 2024.

\bibitem[Lempitsky et~al.(2009)Lempitsky, Kohli, Rother, and
  Sharp]{lempitsky2009image}
Victor Lempitsky, Pushmeet Kohli, Carsten Rother, and Toby Sharp.
\newblock Image segmentation with a bounding box prior.
\newblock In \emph{ICCV}, pages 277--284, 2009.

\bibitem[Liu et~al.(2023)Liu, Xu, Bertasius, and
  Niethammer]{liu2023simpleclick}
Qin Liu, Zhenlin Xu, Gedas Bertasius, and Marc Niethammer.
\newblock {SimpleClick:} interactive image segmentation with simple vision
  transformers.
\newblock In \emph{ICCV}, pages 22290--22300, 2023.

\bibitem[McGuinness and O’connor(2010)]{mcguinness2010comparative}
Kevin McGuinness and Noel~E O’connor.
\newblock A comparative evaluation of interactive segmentation algorithms.
\newblock \emph{Pattern Recognition}, 43\penalty0 (2):\penalty0 434--444, 2010.

\bibitem[Rother et~al.(2004)Rother, Kolmogorov, and Blake]{rother2004grabcut}
Carsten Rother, Vladimir Kolmogorov, and Andrew Blake.
\newblock {GrabCut:} interactive foreground extraction using iterated graph
  cuts.
\newblock \emph{ACM transactions on graphics (TOG)}, 23\penalty0 (3):\penalty0
  309--314, 2004.

\bibitem[Sofiiuk et~al.(2020)Sofiiuk, Petrov, Barinova, and
  Konushin]{sofiiuk2020f}
Konstantin Sofiiuk, Ilia Petrov, Olga Barinova, and Anton Konushin.
\newblock {f-BRS}: Rethinking backpropagating refinement for interactive
  segmentation.
\newblock In \emph{CVPR}, pages 8623--8632, 2020.

\bibitem[Sofiiuk et~al.(2022)Sofiiuk, Petrov, and
  Konushin]{sofiiuk2022reviving}
Konstantin Sofiiuk, Ilya~A Petrov, and Anton Konushin.
\newblock Reviving iterative training with mask guidance for interactive
  segmentation.
\newblock In \emph{ICIP}, pages 3141--3145, 2022.

\bibitem[Xu et~al.(2016)Xu, Price, Cohen, Yang, and Huang]{xu2016deep}
Ning Xu, Brian Price, Scott Cohen, Jimei Yang, and Thomas~S Huang.
\newblock Deep interactive object selection.
\newblock In \emph{CVPR}, pages 373--381, 2016.

\end{thebibliography}
}

\end{document}